\documentclass[preprint,12pt,authoryear]{elsarticle}
\graphicspath{ {./figures/} }
\usepackage{hyperref}
\usepackage{float}
\usepackage{verbatim} 
\usepackage{apalike}
\restylefloat{figure}
\restylefloat{table}
\usepackage{appendix}
\usepackage{graphicx}
\usepackage{subfig}
\usepackage{xcolor}
\usepackage{booktabs} 
\usepackage{amsmath}
\usepackage{multirow}
\usepackage{geometry}
\usepackage{xcolor}
\usepackage{hhline}

\journal{Machine Learning with Applications}

\begin{document}
\begin{frontmatter}

\begin{titlepage}
\begin{center}
\vspace*{1cm}

\textbf{ \large SBERT-InterpEVAL: A Semantic Evaluation Framework for Large Language Models in Fine-Grained Mental Health Topic Classification}

\vspace{1.5cm}

Cristina Luna-Jiménez$^{a,b}$ (cristina.luna.jimene@uni-a.de), 
Nicolas Wagner $^{d}$ (nicolas.wagner@uni-bamberg.de), 
Stefan Ultes $^d$ (stefan.ultes@uni-bamberg.de), 
Elisabeth André $^b$ (elisabeth.andre@uni-a.de), 
David Griol$^a$ (dgriol@ugr.es), 
Zoraida Callejas$^{a,c}$ (zoraida@ugr.es)
\\

\hspace{10pt}

\begin{flushleft}
\small  

$^a$ Departamento de Lenguajes y Sistemas Informáticos, E.T.S. de Ingenierías Informática y de Telecomunicación, Universidad de Granada, C/ Periodista Daniel Saucedo Aranda S/N, 18071, Granada, Spain.

$^b$ Chair for Human-Centered Artificial Intelligence, Faculty of Applied Computer Science, University of Augsburg, Universitätsstraße 2, 86159 Augsburg, Germany.

$^c$ Centro de Investigación en Tecnologías de la Información y las Comunicaciones (CITIC-UGR), Granada, Spain.

$^d$ Natural Language Generation and Dialogue Systems Group, University of Bamberg, Kapuzinerstraße 16, 96047 Bamberg, Germany.


\vspace{1cm}
\textbf{Corresponding Author:} \\
Cristina Luna-Jiménez\\
Chair for Human-Centered Artificial Intelligence, Faculty of Applied Computer Science, University of Augsburg, Universitätsstraße 2, 86159 Augsburg, Germany.
Tel: (+49) 821 598 2348  \\
Email: cristina.luna.jimenez@uni-a.de \\

\end{flushleft}        
\end{center}

\end{titlepage}

\title{SBERT-InterpEVAL: A Semantic Evaluation Framework for Large Language Models in Fine-Grained Mental Health Topic Classification}

\author[label1,label2]{Cristina Luna-Jiménez \corref{cor1}}
\ead{cristina.luna.jimenez@uni-a.de}

\author[label4]{Nicolas Wagner}
\ead{nicolas.wagner@uni-bamberg.de}

\author[label4]{Stefan Ultes}
\ead{stefan.ultes@uni-bamberg.de}

\author[label2]{Elisabeth André}
\ead{elisabeth.andre@uni-a.de}

\author[label1]{David Griol}
\ead{dgriol@ugr.es}

\author[label1,label3]{Zoraida Callejas}
\ead{zoraida@ugr.es}

\cortext[cor1]{Corresponding author.}
\address[label1]{Departamento de Lenguajes y Sistemas Informáticos, E.T.S. de Ingenierías Informática y de Telecomunicación, Universidad de Granada, C/ Periodista Daniel Saucedo Aranda S/N, 18071, Granada, Spain}
\address[label2]{Chair for Human-Centered Artificial Intelligence, Faculty of Applied Computer Science, University of Augsburg, Universitätsstraße 2, 86159 Augsburg, Germany.}
\address[label3]{Centro de Investigación en Tecnologías de la Información y las Comunicaciones (CITIC-UGR), Granada, Spain}
\address[label4]{Natural Language Generation and Dialogue Systems Group, University of Bamberg, Kapuzinerstraße 16, 96047 Bamberg, Germany.}

\begin{abstract}
It is estimated that 46\% of EU citizens experienced psychosocial issues in 2023, highlighting the need for scalable tools to support mental healthcare. Natural language processing (NLP) can assist clinicians by automatically analyzing users’ written expressions, enabling diagnosis and longitudinal monitoring. While prior research has focused on fine-tuning language models to identify specific mental health conditions (e.g., anxiety or depression), it remains unclear whether domain-adapted models generalize to fine-grained classification of diverse mental health topics, including conditions, causes, and symptoms, compared with their general-domain counterparts. This challenge is compounded by the scarcity of richly annotated datasets and the lack of fair evaluation methodologies across different language model architectures. To address these gaps, we evaluate two conversational datasets comprising more than 28 mental health topics and introduce a semantic evaluation framework that enables equitable zero-shot comparison between masked (RoBERTa) and causal (Llama-2) language models by projecting model outputs and topic labels into a shared embedding space. We further compare general-purpose and mental-health-adapted variants using both zero-shot prompting and embedding-based classification. Our results reveal that open-domain models can unexpectedly outperform domain-adapted models under zero-shot evaluation, whereas this advantage diminishes or reverses when using embedding extraction. Additionally, causal language models generally outperform masked language models in topic
recognition, albeit at greater computational cost. These findings demonstrate that conclusions about domain adaptation depend strongly on the evaluation strategy and highlight the importance of employing complementary evaluation methods when assessing language models for fine-grained mental health topic classification.
\end{abstract}

\begin{keyword}
Natural Language Processing \sep Large Language Models \sep Evaluation \sep Topic classification \sep Zero-Shot
\end{keyword}

\end{frontmatter}

\section{Introduction}
\label{introduction}

According to the World Health Organisation, around one in eight people in the world live with a mental disorder. However, most people suffering from mental health conditions do not have access to effective care because of different reasons: unavailability of services or support, lack of capacity of health systems, inaccessibility, or non-affordable costs. On top of this, the widespread stigma associated with mental health disease is still a barrier that stops people from seeking help' \cite{WHO_mentalHealth}. In these scenarios, conversational agents and technological solutions have been proposed as a complement to help users to record their historic, thoughts and living experiences, which could offer support to expert therapists for an early-diagnosis and track user state evolution. 

Due to the vulnerability of this population group, most chatbots follow rule-based approaches to offer higher control over the answers provided \cite{Callejas2021review}. This is the case of Woebot \cite{woebot}, Wysa \cite{wysa}, ChatPal \cite{chatPAL}, and other assistants \cite{Kraus2021} that have been carefully handcrafted using expert knowledge. However, these solutions may struggle with understanding user queries outside the scope of their predefined intents, or with processing nuanced expressions.

In contrast, Large Language Models (LLMs) open a new paradigm, since they do not follow predefined flows and, thus, they can offer more flexible interactions. This specific feature is specially relevant in Natural Language Processing applied to mental-health, since users could express and define their ideas, questions and thoughts with the help of the LLM, which is something not-feasible with rule-based chatbots. However, LLMs have the disadvantage of being less predictable and their answers depend highly on massive training datasets, normally not supervised or filtered by human specialists. For this reason, studying their capacities, limitations and propose new methods to evaluate their responses \cite{MentalHealthAgent,cbms24} is highly important to avoid negative consequences, especially for these sensitive domain. 

In this paper, we focus on a relevant aspect: their capability to distinguish between different topics of conversation related to mental health, since having this capability is fundamental to determining the coherence of the system's responses provided the user input in the current context \cite{RodrguezCantelar2023}. With this aim, we evaluate two families of Language Models with different: 1) architectures, RoBERTa vs. LLama-2 ;2) sizes in terms of the number of parameters, 124MB versus 7B of parameters; and 3) learned knowledge and intended domains of application, since ones are pre-trained for being exploited in open-domains, and others for mental-health scenarios. 


To perform the evaluation and comparison between Language Models, we needed to select large and realistic datasets annotated in terms of mental health. However, as this domain is quite specific and constrained, the number of available datasets is reduced, and most of them are collected from open forums such as Reddit \cite{CLPsych19}, or Twitter \cite{CLEFconf}, which may contain inappropriate responses. Moreover, they also lack information about the topic (or sub-topics) of the dialogues or posts, even though this is a significant factor that can provide LLMs context to generate accurate responses. For this reason, we provide the software to acquire a novel dataset to study mental health disorders and evaluate LLMs in mental health topic classification. Unlike other datasets, this corpus contains questions and answers moderated by expert therapists. Additionally, it is one of the largest datasets in the domain with more than 142,000 samples captured in a real-world scenario. 



Apart from the contribution of the dataset, by evaluating cutting-edge LLMs we also address the following research question: 

\begin{itemize}
    \item \textbf{RQ1:} Do pre-trained models on mental health perform well in categorizing mental health topics? Or do their general-domain versions outperform them? This question would allow us to understand the generalization capacity of these models and to decide whether future efforts should be dedicated to improving the open-domain models or, instead, their adapted versions to mental health. 
    \item \textbf{RQ2:} Could different architectures of LLMs (e.g. LLama-2 vs. RoBERTa) present differences in topic recognition in the mental-health domain? This question sheds light on the balance between the number of parameters and performance in these two architectures since more computational resources are required for LLama than for RoBERTa, which could be a limiting factor in certain scenarios. 
    \item \textbf{RQ3:} How can we compare different architectures of LLMs at Zero-Shot when their outputs have different structures? This question arises from a possible gap in the literature since existing Zero-Shot approaches are not equivalent in Causal Language Models (e.g. Llama) and Masked Language Models (e.g. RoBERTa); hence, their comparisons are neither fair. For this reason, we propose the SBERT-InterpEval methodology oriented to compare Zero-Shot responses in classification tasks for both types of models.
    
\end{itemize}

The remainder of the paper is organised as follows. Section \ref{sec:relatedWorks} summarises previous related work on available data and models, with a special focus on the field of mental health. Next, Section \ref{sec:methodology} describes the methodology, including data acquisition, and the selection of proposed approaches to carry out topic classification. In Section \ref{sec:setup}, the experimental setup is provided. This is followed by a discussion of the main results obtained in Section \ref{sec:results}. Finally, section \ref{sec:conclusions} summarises the main conclusions of the research carried out and proposes future lines of research.

\section{Related Works}\label{relatedWorks}
\label{sec:relatedWorks}


In this section, we will first summarize several of the available datasets related to mental health; second, we will describe novel models pre-trained on this domain; and third, we will conclude with previous work on topic classification. 

\subsection{Datasets on Mental-Health}
In the literature, many of the existing datasets for performing topic classification on mental health disorders employ data collected from forums, since these forums act as real-world scenarios in which people suffering from these distresses can express their concerns, without compromising their anonymity. This is the case of, for example, SchizzoSQUAD Dataset \cite{squizofreniaQA}, a dataset aimed at training chatbots to learn to answer questions about schizophrenia.

Other released datasets classify mental disorders from posts, such as eRISK, \cite{CLEFconf,CLPsych19,LTEDI2022}. Specifically, one of the novelties of the eRISK dataset \cite{CLEFconf} is its focus on early diagnosis of depression by tracking linguistics expressions employed in posts from Twitter, MTV’s A Thin Line, and Reddit.
CLPsych'19 DS \cite{CLPsych19} follows a similar annotation procedure to the one employed in eRISK based on discovering explicit user statements on their psychological condition (e.g. ``In 2013, I was diagnosed with depression'') to annotate the data. Unlike eRISK, CLPsych'19 DS focuses on distinguishing between different levels of suicidal risks (users at no, low, moderate, or severe risk) in Reddit posts. Regarding the contribution on datasets of Sampath,  and Durairaj \cite{LTEDI2022}, they released a corpus also collected from forums to classify levels of depression (not, moderated and severe). Other datasets with depression and suicide risk assessments extracted from forums include CAMS \cite{CAMS} or SenseMood \cite{SenseMood}.

However, one of the main limitations of most datasets is that they probably could not been employed to train conversational agents in Q/A or for maintaining counseling dialogues with people who are likely suffering from mental health disorders because none of them were captured under the supervision of a therapist that can evaluate the quality of the answers, neither whether the tone of the replies could be adequate as for being considered a clinical response. When we search for this requirement, the list of available corpus decreases drastically. Some available corpus containing expert knowledge are DAIC-WOZ\cite{daicWOz} and HOPE-DS\cite{hopeDS}, with dialogues between patients and therapists in a counseling environment. The first is oriented to depression diagnosis, and the second to general counseling. Another interesting dataset is MotiVAte-DS\cite{memoDS} with synthetic dialogues built upon excepts of the posts extracted from forums' threads. From this list, the most related to our task is Counsel-Chat \cite{bertagnolli2020counsel}, which incorporates pairs of Questions-Answers with the associated topic of the query (e.g. anxiety, depression sleep-improvement, etc.), incorporating answers provided by therapists and professionals in mental health. Nonetheless, one of the main limitations of this last dataset is that it only contains 2,775 pairs of question-answers, with 940 unique questions.

\subsection{Transformer-based Models on Mental-Health}

Regarding models, as a consequence of the evolution of Large Language Models trained for solving general tasks, researchers are starting to evaluate their performance in distinct domains to explore their capabilities and limitations. To simplify the different families of LMs that exist, we classify them into the two considered in this article, which present variations in their architectures, manner of processing inputs, and abilities to solve different tasks: 

\begin{itemize}
    \item Causal Language Models (CLM): These models were developed to predict the next token based on past information, being unidirectional. Some examples are GPT \cite{gpt3} or Llama \cite{llama2} models. These LLMs have demonstrated remarkable performance in generating large chains of text. For this reason, alternative fine-tuned versions adapted to diverse domains are emerging, such as MentaLlama\cite{yang2023mentalllama} which is a Llama-2 model aimed at solving mental-health-related questions. Due to its relatively recent release and the datasets employed for training MentaLlama, this model still needs to be evaluated in real-world scenarios to understand their possibilities. 
    
    \item Masked Language Models (MLM): Unlike CLM, MLMs receive a sequence of topics with masked tokens and the task is to predict which tokens fit in the masked positions. Furthermore, these models are bidirectional, hence, they can pay attention to a more general context, and although they could be employed to generate sequences of text, their performance is better in classification tasks (e.g. sentiment or emotion prediction \cite{bertSentiment,BERTemotions}). BERT is probably the most illustrative sample of this family \cite{BERT}. Regarding mental health applications, a collection of BERT-based models has been recently proposed, achieving state-of-the-art rates in mental-health disorders classification \cite{mentalbert}, from which Mental-RoBERTa seems the most promising model.
\end{itemize}

It is important to remark that both architectures derive from the original transformer \cite{transformer}, but considering only the decoder (CLM) or the encoder (MLM).

As commented in the introduction, these differences in architectures and their output structures make their comparison a challenging task, therefore, we propose an assimilation strategy to compare their outputs at Zero-Shot scenarios called SBERT-InterpEVAL. This novel method is inspired by the previous work of S. Sarkar et al. \cite{sarkar-etal-2023-zero}, in which they perform topic classification comparing several sentence encoders (SBERT\cite{sbert}, USE, LASER, and InferSent) to discover the topics associated with an article. Among their strategies, the simplest that achieved the best results consisted of 1) feeding the keywords of the topics to a sentence encoder (e.g. SBERT \cite{sbert}) to generate an embedding vector($E_{T_i}$); 2) feeding the document to classify by the same encoder to generate an embedding vector ($E_D$); 3) calculating pairwise cosine similarity between the embeddings of the topics ($E_{T_i}, \quad \textrm{where} \quad i=\{0,1,...N\}$, being N the total number of topics) and the embeddings of the article ($E_D$) and; 4) assigning the closest topics to the article. In our strategy, we substituted step 2) by feeding the sentence encoder with the responses of the Language Models to a prompt, projecting their answers into a common and comparable vector space. Apart from the possibility of performing an apple-to-apple comparison between different families of models (CLMs with MLMs), this simple methodology also permits generating confusion matrices to analyze errors committed between classes and recover responses with similar semantic meaning (e.g. when the LM says `anger' and the topic is `anger-management') counting them as correct answers instead of errors, contributing to obtain more realistic and accurate measurements of the Language Models' performance on the classification task.

\subsection{Topic Classification}
Transformers have been employed in many tasks, including topic classification \cite{bertTopic,bertTopicArabic,cbms24}. From the perspective of Conversational Agents, detecting the topic of a conversation is key since this is the first step for solving questions and participating effectively in an interaction. For this reason, topic classification has been extensively explored, for example, as part of argumentative systems \cite{carolineArgSys}, or as part of an inconsistencies detector in dialogues \cite{RodrguezCantelar2023}. 

Although several methods exist for classifying topics, two of the most successful in terms of scalability and real-time performance are Zero-Shot and Word Embedding. These methods exploit pre-trained models to perform topic classification even when the initial set of topics is different from the new target set, enabling increasing the initial space of topics or detecting outliers inside each topic \cite{zeroShotTopic}. In this regard, SBERT \cite{sbert}is one of the most employed sentence encoders since it was trained following the triplet-loss approach \cite{facenet}. This loss function minimizes the distance between sentences with similar semantic meanings while increasing the distance between more dissimilar words in an Euclidean space. As a result, the embeddings generated by this model contain relevant semantic information, related to the topic of the sentence. For this reason, these embeddings are frequently used for discovering clusters of new topics and sub-topics or in classification tasks \cite{RodrguezCantelar2023,memText,topicsModel}.

Moreover, Zero-Shot can also be directly applied to explore the capabilities of LLMs, since they can receive and generate large chains of texts, as in the approaches followed in \cite{yang2023mentalllama,yang-etal-2023-towards,sarkar-etal-2023-zero} to classify depression, mental health issues or other topics belonging to different domains. Extending this Zero-Shot technique to topic classification could give clues about the capacity of the LMs to handle vocabulary related to more specific domains, such as mental health.




\section{Materials and Methods}
\label{sec:methodology}
This section explains the data collection and the approaches for comparing domains and architectures to solve fine-grained mental health topic classification. 

\subsection{Data collection}\label{sec:dataCollection}

We employ two datasets, `Counsel-Chat' and `7Cups' since both are annotated in mental health topics and follow the same structure of Questions/Answers. 

\begin{figure}[h]
    \centering
    \includegraphics[width=0.7\linewidth]{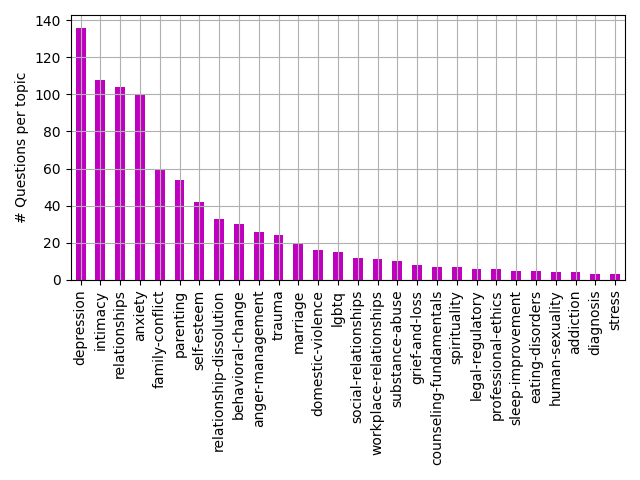}
    \caption{Overview of the number of questions per topic in the Counsel-Chat dataset, ranging from 1 to 136.}%
\label{fig:topicsCounselChat}%
\end{figure}

The counsel-chat corpus is publicly available in the HuggingFace repository (`nbertagnolli/counsel-chat') and contains questions and answers extracted originally from `www.counselchat.com'. This platform gathers the questions from users who posted their mental health issues, and the answers provided by professional therapists. The dataset contains 2,775 question-answer pairs, with 863 unique questions belonging to 30 unbalanced topics containing diseases (e.g. depression, anxiety, eating disorders, etc.), symptoms (e.g. anger management), and possible causes (e.g. relationships). The distribution of classes is plotted in Figure \ref{fig:topicsCounselChat}. 

\begin{figure}[H]
    \centering
    \includegraphics[width=0.7\linewidth]{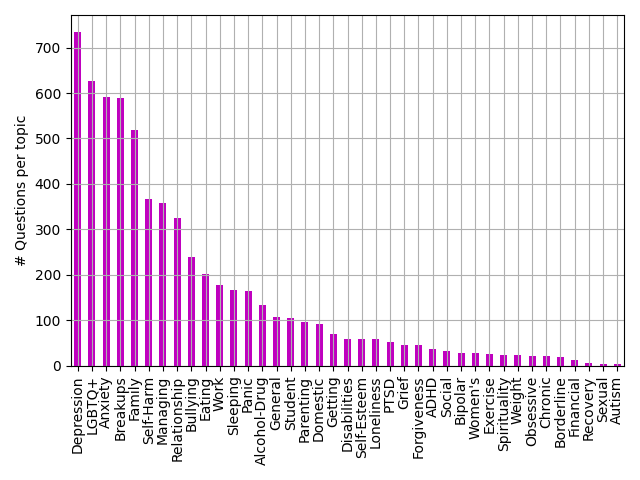}
    \caption{Overview of the number of questions per topic in the 7Cups dataset, ranging from 3 to 735.}%
\label{fig:topics7cups}%
\end{figure}

Unlike Counsel-Chat, the `7cups dataset' was collected from the `https://www.7cups.com/qa' website. This platform follows a similar idea to the Counsel-Chat; each forum is assigned to an expert in charge of moderating it. In addition to the users' votes, the platform allows experts to vote, differentiating their votes from those provided by regular users (called `listeners'). These listeners are also distinguished in proficiency levels by topic, assigned in concordance with their contributions and the quality of their responses. The web-scrapping process resulted in 136,102 samples with 6,270 unique questions on 39 topics, ranging from 3 to 735 questions per topic. The distribution of classes is plotted in Figure \ref{fig:topics7cups}.

\subsection{Topic Classification: Zero-Shot Approaches}\label{sec:zeroShot}

First, to analyze the first research question (RQ1), our initial hypothesis is that expert systems more adapted to a specific domain should perform better in tasks related to that domain. 

In our experiments, the scenario to study is mental health, with the previously mentioned datasets, counsel-chat and 7cups, since both contain inquiries posted by users who might suffer from a mental health disorder annotated by topic. 

To solve the research question, we compare open-domain pre-trained models with their corresponding versions adapted to the mental health domain, to understand whether we could verify or deny the hypothesis for our use-case.


Second, to address RQ2 and explore differences across families of models at Zero-Shot approaches, we select RoBERTa and Llama models. As these LMs differ in architectures, size of input context and format of their responses, we need to design and adapt the prompts to each model. Specifically, as Llama-2 accepts a longer context (up to 4,096 tokens) at its inputs than RoBERTa (512 tokens), the prompt of Llama-2 includes the post plus the complete list of topics of each dataset, as depicted in Table \ref{tab:promptLLamaZeroShot}. 

\begin{table}[h] 
  \caption{Prompt template for classifying topics with (Mental-)Llama-2}
  \label{tab:promptLLamaZeroShot}
  \centering
  \begin{tabular}{l}
    \hline 
    \textbf{Prompt:} \\
    Consider this post: ``Anger often comes with a lot of other emotions. \\
    Consider what else you may be feeling at the same time and keep \\
    track of your different mood changes. You may be able to \\
    notice patterns, such as being more frustrated when you don't sleep \\
    well. If you do feel as though you are mad all the time, can you think \\
    of any exceptions so you can see what is different in those times?''. \\
    {[}Question: Which topic does the text belong to? Select one from the list: \\
    depression, anxiety, trauma, anger-management, ... {]} \\
    \hline 
    \textbf{Response:} \\
    The text belongs to the topic of `anger-management'.\\
    \hline
  \end{tabular}
\end{table}

\begin{table}[h] 
  \caption{Prompt template for classifying topic with (Mental-)RoBERTa}
  \label{tab:promptBERTZeroShot}
  \centering
  \begin{tabular}{l}
    \hline
    \textbf{Prompt:} \\
    ``Anger often comes with a lot of other emotions. \\
    Consider what else you may be feeling at the same time and keep \\
    track of your different mood changes. You may be able to \\
    notice patterns, such as being more frustrated when you don't sleep \\
    well. If you do feel as though you are mad all the time, can you think \\
    of any exceptions so you can see what is different in those times?''. \\
    The topic of the text is \textless{}mask\textgreater{}.\\
    \hline
    \textbf{Response:} \\
    anger \\
    \hline
  \end{tabular}
\end{table}

Unlike Llama-2, RoBERTa prompts only contain the posted question (or answer) together with the sentence 'The topic of the text is $\textless{}mask\textgreater{}$', as appears in Table \ref{tab:promptBERTZeroShot}, with the aim that it completes the masked token with the topic of the post. 

Apart from the distinct length of the context, the models' responses also follow different structures. Specifically, RoBERTa returns the 5-top most probable words that can fit in the masked token position, providing structured and easy-to-recover answers. In contrast, Llama models do not always return a structured pattern e.g. in the response there might appear a discussion about the suitability of choosing one topic or another as part of the reasoning process of the LLM; consequently, several topics may be present in the same response. From this observation, we reach the RQ3, to study how we could compare different architectures of LLMs at Zero-Shot when their outputs have different structures. Thereupon, the SBERT-InterpEval strategy is proposed to cope with their differences. As this is a novel strategy, we compare it with other traditional Zero-Shot approaches in the literature to evaluate its performance.

\begin{figure*}[htpb]
\centering
    \caption{Illustration of the SBERT-InterprEVAL approach for error analysis of the topics predicted by RoBERTa and Llama models on Zero-Shot approaches.}
\label{fig:semSim}
    \includegraphics[width=0.79\textwidth]{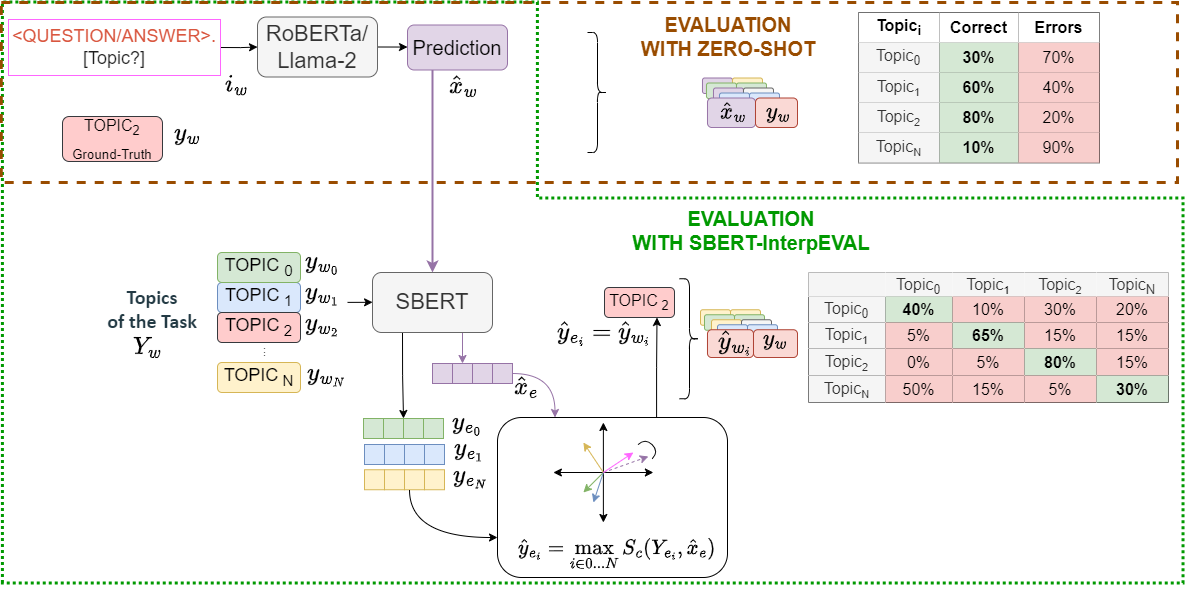}
\end{figure*}

In detail, the complete list of Zero-Shot strategies that we use for assessing the performance of RoBERTa are:
\begin{itemize}
    \item Zero-Shot (on Top-1): In this strategy, only the first word returned by the model is considered as the prediction of the topic to fill in the masked token. When this predicted word matches the ground truth, it is counted as a correct answer; otherwise, it is counted as an error. 
    \item Zero-Shot (on Top-5): This strategy is a variation of the Zero-Shot (on Top-1) version, in which a prediction is considered correct if the ground truth matches with at least one of the top-5 words provided by the model for the masked token. This metric is more grateful than Zero-Shot (on Top-1) since judgments are based on the 5-top answers instead of just on 1; nevertheless, it is frequently used for reporting results in classification tasks with many classes \cite{xception}. 
    \item SBERT-InterpEVAL: We propose this strategy that considers the cosine similarity between the top-1 most probable word and all the possible topics' keywords. As it is illustrated in Figure \ref{fig:semSim}, the procedure is divided into the following steps: 1) applying the Zero-Shot (on Top-1) strategy to produce the responses of the LM ($\hat{x_w}$), 2) feeding a sentence encoder (SBERT in our case) with this response to generate an embedding vector ($\hat{x_e}$), 3) feeding the same sentence encoder (SBERT) with the names of the topics ($\hat{y_wi}$) to generate one embedding per topic ($\hat{y_ei}$), 4) calculating pairwise cosine similarity between the embeddings of the topics ($\hat{y_wi}$, where i = {0, 1, ...N }, being N the total number of topics) and the embedding of the LM prediction ($\hat{x_e}$), 5) assigning the closest topic ($\hat{y_wi}$) to the prediction and compare with the ground-truth to estimate the performance of the model on topic classification (${y_w}$). The main advantage of this strategy is that it can solve lemmatization problems (e.g. when the top-1 word is `problem sleeping' and the ground truth is `sleep-problems', previous strategies considered the LM prediction as an error, but in this case, SBERT-InterpEVAL count it as a success). Additionally, it homogenizes conditions with Llama by providing the additional topics not passed in the prompt of RoBERTa due to lack of context, which facilitates the ensemble of different LMs \cite{iwinac2024}. Finally, it also allows detailed analysis of errors across topics, contrary to other Zero-Shot approaches that only account for errors inside the same topic, without supplying information about the most confused classes with each specific topic to study their differences or similarities. 
\end{itemize}

About the Zero-Shot strategies to evaluate Llama, we select two alternatives employed in the literature to contrast with SBERT-InterpEval method:

\begin{itemize}
\item Zero-Shot: As commented before, Llama can provide unstructured responses with several topics. This Zero-Shot strategy considers successful responses when the predicted topic matches the ground truth and no other topics occur in the same response. In opposition to this, all the responses with more than one topic count as errors, even though the ground truth is included in them.

\item N-Shots: This approach is less restrictive than the first Zero-Shot strategy of Llama, considering as errors only those cases in which the ground truth does not appear in the response. 
 
 \item SBERT-InterpEVAL: This strategy is applied similarly to how it was explained for RoBERTa, but this time $\hat{x_w}$ is the unstructured response of Llama. Apart from the previously mentioned advantages, another benefit for Llama is that this procedure permits selecting a unique topic from the unstructured response of the LLM, giving it a structure and warranting the application of classification metrics.

\end{itemize}

\subsection{Topic Classification: Fine-Tuning Approaches}\label{sec:topicClassif}

Alternatively, to the Zero-Shot approaches and to explore further RQ1 and RQ2, we extract embeddings from the last hidden state of RoBERTa and LLama-2 models. Subsequently, we calculate the average across the temporal dimension to obtain a compact representation of the fed sequence, resulting in a final 768-dimensional embedding for RoBERTa and a 4,096-dimensional embedding for LLama-2. 

Additionally, we train a Continuous Bag of Words (CBOW) model to encode words into dense real-value vector representations to compare RoBERTa and LLama embeddings with this standard strategies in the field. Likewise, we extract the embeddings of the SBERT sentence encoder, which demonstrated state-of-the-art performance in topic classification and topic discovering tasks, as a competitive baseline to measure disparities with the RoBERTa and Llama embeddings. It is important to mention that unlike SBERT encoder, which was pre-trained to differentiate between similar and dissimilar sentences contrasting pairs of them, Llama embeddings are not customized to the task since this model was originally pre-trained to act as a chatbot to maintain conversations. Nevertheless, it is relevant to evaluate whether its capacities can extend to other tasks, such as topic classification, and how they perform compared to traditional and state-of-the-art solutions. 


Once embeddings of different models are generated, we train several classifiers (SVM, k-NN, NC, RF and MLP) with diverse settings to analyze which vectors achieve the highest performance on topic classification related to mental health.

\section{Experimental Set-Up}\label{sec:setup}

\subsection{Datasets}
As we commented before in Section \ref{sec:dataCollection}, we employ the `Counsel-Chat' dataset from HuggingFace library \cite{bertagnolli2020counsel}, and the `7Cups' dataset for performing mental-health topic classification. Although we used most of the samples of the dataset, we filtered those topics that did not contain at least 3 questions, as well as the repeated samples (some answers were duplicated for different questions). Finally, we split the pairs of questions and answers into independent samples annotated by topic to perform topic classification on them. After applying these filters, the Counsel-Chat dataset contains 3,451 samples in total, belonging to 28 different topics; while 7cups contained 142,230 samples on 39 different topics. To evaluate our methods we divide the complete sets of samples into train, validation, and test sets in an 80/10/10 proportion. In the end, Counsel-Chat contains 2,545/419/487 samples and 7-Cups 109,076/15,466/17,688 samples per training/validation/test sets.

\subsection{Models}
We select RoBERTa (`roberta-base') and Llama-2 (`meta-llama/Llama-2-7b-chat-hf') as the open-access solutions because they have pre-trained versions adapted to mental-health (`mental/mental-roberta-base' and `klyang/MentaLlama-chat-7B'), which are appropriate to solve all the research questions of this article. Although Llama is most commonly used for next-token prediction, previous studies have adapted them for reasoning in prompts to return classes \cite{yang2023mentalllama}; hence, we consider them suitable to be explored for mental-health topics' classification using the Zero-Shot and embedding extraction methods.




For training the CBOW baseline model, we select a vector size equal to 768 for projecting words into this new space and be comparable with the embeddings of RoBERTa, the rest of the parameters were maintained to their default values in the Gensim library. The embeddings of the Siamese-BERT network are extracted from `sentence-transformers/all-mpnet-base-v2' \cite{sbert} implementation available in HuggingFace. 

After extracting these embeddings, we train distinct classifiers available in the sklearn library for solving topic classification. The specific combinations of models and parameters that we evaluate are: SVM with `RBF' or `linear' kernel and C ranging from 0.001 to 100 in steps multiple of ten (0.001,0.01,0.1,1.0,10,100); logistic regression models with C ranging from 0.001 to 10 in steps multiple of 10; k-NN with neighbors ranging from 5 to 30 in steps of 5 (5,10,15,20,25,30); Nearest Centroid (NC) algorithm with default parameters; Decision Trees with min\_samples\_split parameter ranging from 5 to 30 in steps of 5; Random Forests (RF) with different n\_estimators ranging from 10 to 90 in steps of 10; and MLPs with 1, 2 or 3 layers of 80 neurons each. 

\subsection{Metrics}
Given the unbalance between samples for each topic and the high number of topics, we report the Accuracy (ACC), Weighted-F1 (W-F1) and Macro-F1 (M-F1) metrics with their 95\% confidence intervals on the test set. However, decisions on the optimal parameters of the top models are made on the development set for the combination that achieves the best W-F1.

\section{Results and Discussion}\label{experiments}
\label{sec:results}

 Tables \ref{tab:ZeroShotTopic} and \ref{tab:ZeroShotTopic7cups} encompass the results obtained from the Language Models and strategies presented before for topic classification in the mental health domain. In what follows, we will discuss the outcomes in context with the research questions that this article addresses. 


\vspace{-2pt}
\subsection{RQ1. Comparison between open-domain models vs. mental-health adapted models}

Regarding differences between domains, it is observable that counter-intuitively in some cases the model most adapted to the mental-health topic detection task (Mental-RoBERTa) performs worse than the general domain versions (RoBERTa) in the Zero-Shot approaches. Examining the most repeated responses for the Zero-Shot (Top-1) strategy of RoBERTa-family models in Table \ref{tab:ZeroShotTopic_top1}, it is noticeable that in the 70\% of the cases for Counsel-Chat and 71.31\% for 7Cups, Mental-RoBERTa predicted for the $<mask>$ token non-concrete words such as `this', `irrelevant' or `important', which partially explain their performance gap. Additionally, its diversity in the vocabulary of predicted topics for the `mask' token is limited compared to the open-domain RoBERTa model. In particular, for the Counsel-Chat dataset, Mental-RoBERTa proposed only 85 different words for topics, whereas the open-domain version of RoBERTa generated 242 terms. A similar pattern can be detected in the 7Cups dataset: Mental-RoBERTa reported just 517 genuine predictions for topics and the open-domain RoBERTa more than 1,800.

\begin{table}[htpb]
\centering
\fontsize{7pt}{6pt}\selectfont
\addtolength{\tabcolsep}{-5pt}
\caption{Top-15 most repeated topics for Zero-Shot (top-1) approach.}
\label{tab:ZeroShotTopic_top1}
\begin{tabular}{|cc|cc|cc|cc|} 
\hline
\multicolumn{4}{|c|}{\textbf{Counsel-Chat (3,451 samples)}}                                                                                                                                                                                                                                                                              & \multicolumn{4}{c|}{\textbf{7Cups (142,230 samples)}}                                                                                                                                                                                                                                                                                        \\ 
\hline
\multicolumn{2}{|c|}{\textbf{RoBERTa}}                                                                                                                              & \multicolumn{2}{c|}{\begin{tabular}[c]{@{}c@{}}\textbf{Mental}\\\textbf{ RoBERTa}\end{tabular}}                                                                    & \multicolumn{2}{c|}{\textbf{RoBERTa}}                                                                                                                                 & \multicolumn{2}{c|}{\begin{tabular}[c]{@{}c@{}}\textbf{Mental}\\\textbf{ RoBERTa}\end{tabular}}                                                                      \\ 
\hline
\begin{tabular}[c]{@{}c@{}}\textbf{Top}\\\textbf{ repeated}\\\textbf{ words}\end{tabular} & \textbf{Freq.}                                                          & \begin{tabular}[c]{@{}c@{}}\textbf{Top}\\\textbf{ repeated}\\\textbf{ words}\end{tabular} & \textbf{Freq.}                                                         & \begin{tabular}[c]{@{}c@{}}\textbf{Top}\\\textbf{ repeated}\\\textbf{ words}\end{tabular} & \textbf{Freq.}                                                            & \begin{tabular}[c]{@{}c@{}}\textbf{Top}\\\textbf{ repeated}\\\textbf{ words}\end{tabular} & \textbf{Freq.}                                                           \\ 
\hline
depression                                                                                & \begin{tabular}[c]{@{}c@{}}631\\ (18.28\%)\end{tabular}                 & \textbf{this}                                                                             & \begin{tabular}[c]{@{}c@{}}1,770\\ (51.29\%)\end{tabular}              & depression                                                                                & \begin{tabular}[c]{@{}c@{}}32,418\\ (22.19\%)\end{tabular}                & \textbf{this}                                                                             & \begin{tabular}[c]{@{}c@{}}84,386\\ (59.33\%)\end{tabular}               \\ 
\hline
love                                                                                      & \begin{tabular}[c]{@{}c@{}}347\\ (10.06\%)\end{tabular}                 & \textbf{irrelevant}                                                                       & \begin{tabular}[c]{@{}c@{}}389\\ (11.27\%)\end{tabular}                & love                                                                                      & \begin{tabular}[c]{@{}c@{}}21,593\\ (15.18\%)\end{tabular}                & \textbf{irrelevant}                                                                       & \begin{tabular}[c]{@{}c@{}}12,319\\ (8.07\%)\end{tabular}                \\ 
\hline
therapy                                                                                   & \begin{tabular}[c]{@{}c@{}}208\\ (6.03\%)\end{tabular}                  & \textbf{important}                                                                        & \begin{tabular}[c]{@{}c@{}}265\\ (7.68\%)\end{tabular}                 & anxiety                                                                                   & \begin{tabular}[c]{@{}c@{}}7,332\\ (5.15\%)\end{tabular}                  & depression                                                                                & \begin{tabular}[c]{@{}c@{}}11,392\\ (8.01\%)\end{tabular}                \\ 
\hline
suicide                                                                                   & \begin{tabular}[c]{@{}c@{}}164\\ (4.75\%)\end{tabular}                  & suicide                                                                                   & \begin{tabular}[c]{@{}c@{}}207\\ (5.99\%)\end{tabular}                 & suicide                                                                                   & \begin{tabular}[c]{@{}c@{}}4,950\\ (3.48\%)\end{tabular}                  & suicide                                                                                   & \begin{tabular}[c]{@{}c@{}}6,798\\ (4.78\%)\end{tabular}                 \\ 
\hline
rape                                                                                      & \begin{tabular}[c]{@{}c@{}}151\\ (4.38\%)\end{tabular}                  & depression                                                                                & \begin{tabular}[c]{@{}c@{}}190\\ (5.51\%)\end{tabular}                 & \textbf{this}                                                                             & \begin{tabular}[c]{@{}c@{}}4,109\\ (2.89\%)\end{tabular}                  & \textbf{important}                                                                        & \begin{tabular}[c]{@{}c@{}}5,566\\ (3.91\%)\end{tabular}                 \\ 
\hline
anxiety                                                                                   & \begin{tabular}[c]{@{}c@{}}129\\ (3.74\%)\end{tabular}                  & therapy                                                                                   & \begin{tabular}[c]{@{}c@{}}116\\ (3.36\%)\end{tabular}                 & meditation                                                                                & \begin{tabular}[c]{@{}c@{}}3,439\\ (2.42\%)\end{tabular}                  & anxiety                                                                                   & \begin{tabular}[c]{@{}c@{}}4,743\\ (3.33\%)\end{tabular}                 \\ 
\hline
sex                                                                                       & \begin{tabular}[c]{@{}c@{}}115\\ (3.33\%)\end{tabular}                  & anxiety                                                                                   & \begin{tabular}[c]{@{}c@{}}87\\ (2.52\%)\end{tabular}                  & religion                                                                                  & \begin{tabular}[c]{@{}c@{}}3,114\\ (2.19\%)\end{tabular}                  & love                                                                                      & \begin{tabular}[c]{@{}c@{}}2,468\\ (1.73\%)\end{tabular}                 \\ 
\hline
divorce                                                                                   & \begin{tabular}[c]{@{}c@{}}99\\ (2.87\%)\end{tabular}                   & sex                                                                                       & \begin{tabular}[c]{@{}c@{}}58\\ (1.68\%)\end{tabular}                  & rape                                                                                      & \begin{tabular}[c]{@{}c@{}}3,006\\ (2.11\%)\end{tabular}                  & sexuality                                                                                 & \begin{tabular}[c]{@{}c@{}}806\\ (0.06\%)\end{tabular}                   \\ 
\hline
counseiling                                                                               & \begin{tabular}[c]{@{}c@{}}94\\ (2.72\%)\end{tabular}                   & anger                                                                                     & \begin{tabular}[c]{@{}c@{}}32\\ (0.93\%)\end{tabular}                  & bullying                                                                                  & \begin{tabular}[c]{@{}c@{}}2,635\\ (1.85\%)\end{tabular}                  & bullying                                                                                  & \begin{tabular}[c]{@{}c@{}}779\\ (0.05\%)\end{tabular}                   \\ 
\hline
\textbf{here}                                                                             & \begin{tabular}[c]{@{}c@{}}78\\ (2.72\%)\end{tabular}                   & PTSD                                                                                      & \begin{tabular}[c]{@{}c@{}}28\\ (0.81\%)\end{tabular}                  & happiness                                                                                 & \begin{tabular}[c]{@{}c@{}}2,316\\ (1.63\%)\end{tabular}                  & gender                                                                                    & \begin{tabular}[c]{@{}c@{}}678\\ (0.05\%)\end{tabular}                   \\ 
\hline
\textbf{important}                                                                        & \begin{tabular}[c]{@{}c@{}}59\\ (1.71\%)\end{tabular}                   & love                                                                                      & \begin{tabular}[c]{@{}c@{}}25\\ (0.72\%)\end{tabular}                  & \textbf{here}                                                                             & \begin{tabular}[c]{@{}c@{}}1,897\\ (1.33\%)\end{tabular}                  & meditation                                                                                & \begin{tabular}[c]{@{}c@{}}656\\ (0.05\%)\end{tabular}                   \\ 
\hline
anger                                                                                     & \begin{tabular}[c]{@{}c@{}}54\\ (1.56\%)\end{tabular}                   & meditation                                                                                & \begin{tabular}[c]{@{}c@{}}15\\ (0.43\%)\end{tabular}                  & friendship                                                                                & \begin{tabular}[c]{@{}c@{}}1,794\\ (1.26\%)\end{tabular}                  & happiness                                                                                 & \begin{tabular}[c]{@{}c@{}}601\\ (0.04\%)\end{tabular}                   \\ 
\hline
below                                                                                     & \begin{tabular}[c]{@{}c@{}}44\\ (1.27\%)\end{tabular}                   & transgender                                                                               & \begin{tabular}[c]{@{}c@{}}14\\ (0.41\%)\end{tabular}                  & transgender                                                                               & \begin{tabular}[c]{@{}c@{}}1,756\\ (1.23\%)\end{tabular}                  & sex                                                                                       & \begin{tabular}[c]{@{}c@{}}597\\ (0.04\%)\end{tabular}                   \\ 
\hline
trust                                                                                     & \begin{tabular}[c]{@{}c@{}}44\\ (1.27\%)\end{tabular}                   & alcohol                                                                                   & \begin{tabular}[c]{@{}c@{}}14\\ (0.41\%)\end{tabular}                  & sex                                                                                       & \begin{tabular}[c]{@{}c@{}}1,558\\ (1.09\%)\end{tabular}                  & anger                                                                                     & \begin{tabular}[c]{@{}c@{}}543\\ (0.04\%)\end{tabular}                   \\ 
\hline
transgender                                                                               & \begin{tabular}[c]{@{}c@{}}43\\ (1.25\%)\end{tabular}                   & rape                                                                                      & \begin{tabular}[c]{@{}c@{}}14\\ (0.41\%)\end{tabular}                  & sexuality                                                                                 & \begin{tabular}[c]{@{}c@{}}1,515\\ (1.06\%)\end{tabular}                  & addiction                                                                                 & \begin{tabular}[c]{@{}c@{}}520\\ (0.04\%)\end{tabular}                   \\ 
\hline
\textbf{TOTAL}                                                                            & \begin{tabular}[c]{@{}c@{}}\textbf{242}\\\textbf{ (100\%)}\end{tabular} & \textbf{TOTAL}                                                                            & \begin{tabular}[c]{@{}c@{}}\textbf{85}\\\textbf{ (100\%)}\end{tabular} & \textbf{TOTAL}                                                                            & \begin{tabular}[c]{@{}c@{}}\textbf{1,839}\\\textbf{ (100\%)}\end{tabular} & \textbf{TOTAL}                                                                            & \begin{tabular}[c]{@{}c@{}}\textbf{517}\\\textbf{ (100\%)}\end{tabular}  \\
\hline
\end{tabular}
\end{table}

However, for the embedding's extraction strategies, the tendency switches, being Mental-RoBERTa the model achieving higher metrics on the classification, although not in a statistically significant manner.

Regarding the comparison between Llama-2 and MentaLlama, the results are more tied. For example, MentaLlama performs worse in Zero-Shot strategies compared to Llama-2; however, for SBERT-InterpEVAL and N-Shots the performance of both models is statistically comparable in the Counsel-Chat dataset; hence, both seem to be competitive in predicting the mental-health topic of the posts. Similar patterns are observed for the 7Cups dataset, although MentaLlama improves the results of Llama-2 in some decimals for the SBERT-InterpEVAL methodology and surpasses its N-Shot strategy in M-F1 by 4.23 percent points. 
Regarding the embedding extraction methods, both models perform similarly on both datasets since their differences are not statistically significant except for the open-domain Llama-2 model that surpasses MentaLlama in M-F1 by 2.48 statistically significant percentage points on the 7Cups dataset.


\begin{table}[!b]
\centering
\small
\fontsize{7pt}{7pt}\selectfont
\addtolength{\tabcolsep}{-3pt}
\caption{Results on topic classification (28 topics) for the test set (487 samples) of Counsel-Chat dataset. The highest, the best. Acronyms: ACC=Accuracy, CI=Confidence Interval (95\%), W-F1=Weighted-F1, M-F1=Macro-F1}\label{tab:ZeroShotTopic}
\begin{tabular}{ccccccc}
\hline
\textbf{Model}                  & \textbf{Param.} & \textbf{Strategy} & \textbf{FT-Classifier} & \textbf{ACC$\pm$CI} & \textbf{W-F1$\pm$CI} & \textbf{M-F1$\pm$CI}  \\ 
\hline
Zero-Rule                       & - & -  & -                    & 16.22$\pm$3.27   & 4.53$\pm$1.85       & 1.00$\pm$0.88    \\ 
\hline
Word2Vec         &-         &  Emb. Extract. (768 dim.)                                                                     & RF (n =30)                                                                             & 9.24$\pm$2.57   & 7.79$\pm$2.38       & 3.05$\pm$1.53         \\ \hline
SBERT         & 110M                                                                       & \begin{tabular}[c]{@{}c@{}}Emb. Extract. (768 dim.)\end{tabular}  & \begin{tabular}[c]{@{}c@{}}Nearest Centroid\end{tabular}                    & \textbf{52.77$\pm$4.43*}     & \textbf{54.49$\pm$4.42*}      & \textbf{45.41$\pm$4.42*}      \\ \hline
\multirow{4}{*}{RoBERTa}        &\multirow{4}{*}{124M}& Zero-Shot (Top-1)      & -     & 10.27$\pm$2.69     & 15.03$\pm$3.17      & 5.90$\pm$2.09        \\ 
\hhline{~~|-|-|-|-|-}
                                & &  SBERT-InterpEVAL               & - & 30.39$\pm$4.08     & 30.09$\pm$4.07       & 22.22$\pm$3.69  \\ 
\hhline{~~|-|-|-|-|-}
                                & &  Zero-Shot (Top-5)   & -            & 21.36$\pm$3.64                 & 28.02$\pm$3.99                & 12.42$\pm$2.09  \\
\hhline{~~|-|-|-|-|-}
& & \begin{tabular}[c]{@{}c@{}}Emb. Extract. (768 dim.)\end{tabular}  & \begin{tabular}[c]{@{}c@{}}MLP (80,80,80)\end{tabular}                      & 38.81$\pm$4.33     & 37.40$\pm$4.30      & 24.56$\pm$3.82      \\ \hline

\hline
\multirow{4}{*}{Mental-RoBERTa}  &\multirow{4}{*}{124M} & Zero-Shot (Top-1)  & -         & 4.93$\pm$1.92                & 8.04$\pm$2.42                & 3.01$\pm$1.52     \\ 
\hhline{~~|-|-|-|-|-}
                                & & SBERT-InterpEVAL               & - & 17.04$\pm$3.34                 & 17.06$\pm$3.34                & 11.63$\pm$2.85  \\ 
\hhline{~~|-|-|-|-|-}
                                & & Zero-Shot (Top-5)    & -           & 12.53$\pm$2.94               & 16.90$\pm$3.33                & 7.79$\pm$2.38      \\ 
\hhline{~~|-|-|-|-|-} & & 
               \begin{tabular}[c]{@{}c@{}}Emb. Extract. (768 dim.)\end{tabular}  & \begin{tabular}[c]{@{}c@{}}RBF SVC (C=100)\end{tabular}                                                              & 41.07$\pm$4.37   & 39.89$\pm$4.35       & 26.00$\pm$3.90          \\ 

\hline
\multirow{4}{*}{Llama-2}         &\multirow{4}{*}{7B}& Zero-Shot        & -   & 39.22$\pm$4.34                 & \textbf{50.34$\pm$4.44}                 & 28.43$\pm$4.01    \\ 
\hhline{~~|-|-|-|-|-}
                                & &  SBERT-InterpEVAL              & - & \textbf{44.97$\pm$4.42}                 & 44.18$\pm$4.41                & 31.34$\pm$4.12  \\ 
\hhline{~~|-|-|-|-|-}
                                & & N-Shots        & -     & 42.51$\pm$4.39                & \textbf{53.75$\pm$4.43}                & 30.50$\pm$4.09    \\ 
\hhline{~~|-|-|-|-|-}
                                & & \begin{tabular}[c]{@{}c@{}}Emb. Extract. (4096 dim.)\end{tabular}  & \begin{tabular}[c]{@{}c@{}}Lin. SVC (C=0.001)\end{tabular}     & \textbf{48.46$\pm$4.44}     & \textbf{47.43$\pm$4.43}      & 32.36$\pm$4.16      \\
\hline
\multirow{4}{*}{MentaLlama}     &\multirow{4}{*}{7B} & Zero-Shot  & -         & 23.82$\pm$3.78               & 33.95$\pm$4.21                 & 18.20$\pm$3.43     \\ 
\hhline{~~|-|-|-|-|-}
                                & &  SBERT-InterpEVAL      & -         & 38.60$\pm$4.32                 & 38.03$\pm$4.31                & 28.07$\pm$3.99  \\ 
\hhline{~~|-|-|-|-|-}
                                & & N-Shots  &-           & 35.93$\pm$4.26               & \textbf{47.17$\pm$4.43}                & 27.67$\pm$3.97     \\
\hhline{~~|-|-|-|-|-}
                                & & \begin{tabular}[c]{@{}c@{}}Emb. Extract. (4096 dim.)\end{tabular}  & \begin{tabular}[c]{@{}c@{}}Lin. SVC (C=0.01)\end{tabular}     & \textbf{49.90$\pm$4.44}   & \textbf{49.66$\pm$4.44}       & \textbf{36.90$\pm$4.29}      \\
\hline
\end{tabular}
\end{table}

\begin{table*}[htpb]
\centering
\small
\fontsize{7pt}{7pt}\selectfont
\addtolength{\tabcolsep}{-3pt}
\caption{Results on topic classification (39 topics) for the test set (17,688 samples) of the 7Cups dataset. The highest, the best.\\ Acronyms: ACC=Accuracy, W-F1=Weighted-F1, M-F1=Macro-F1, CI=Confidence Interval (95\%). }\label{tab:ZeroShotTopic7cups}
\begin{tabular}{ccccccc}
\hline
\textbf{Model}                  & \textbf{Params.} & \textbf{Strategy} & \textbf{FT-Classifier} & \textbf{ACC$\pm$CI} & \textbf{W-F1$\pm$CI} & \textbf{M-F1$\pm$CI}  \\ 
\hline
Zero-Rule                       & - & - & -                   &14.26$\pm$0.51              & 3.56$\pm$0.27                & 0.64$\pm$0.12   \\ \hline
Word2Vec             &-     & Emb. Extract. (768 dim.)                                                                     & SVC (C=0.1)                                                                             & 53.22$\pm$0.73   & 51.28$\pm$0.74       & 26.69$\pm$0.65         \\ \hline

SBERT       & 110M                                                                         & \begin{tabular}[c]{@{}c@{}}Emb. Extract. (768 dim.)\end{tabular}  & \begin{tabular}[c]{@{}c@{}}RBF -SVC (C=1.0)\end{tabular}                    & \textbf{64.87$\pm$0.70}     & \textbf{63.83$\pm$0.71*}      & 44.38$\pm$0.73      \\ \hline

\multirow{4}{*}{RoBERTa}        &\multirow{4}{*}{124M}& Zero-Shot (Top-1)        & -   & 11.72$\pm$0.47              & 17.25$\pm$0.56                 & 18.29$\pm$0.57    \\ 
\hhline{~~|-|-|-|-|-}
                                & &  SBERT-InterpEVAL   & -             & 26.06$\pm$0.65                 & 25.97$\pm$0.65                & 20.60$\pm$0.60  \\ 
\hhline{~~|-|-|-|-|-}
                                & &  Zero-Shot (Top-5)   & -             & 24.15$\pm$0.63                 & 31.77$\pm$0.69                & 30.43$\pm$0.68  \\ 

\hhline{~~|-|-|-|-|-}
                           & &     \begin{tabular}[c]{@{}c@{}}Emb. Extract. (768 dim.)\end{tabular}  & \begin{tabular}[c]{@{}c@{}} RBF-SVC (C=100)\end{tabular}                      & 59.45$\pm$0.72     & 58.32$\pm$0.73      & 37.88$\pm$0.71      \\

\hline
\multirow{4}{*}{Mental-RoBERTa}  &\multirow{4}{*}{124M} & Zero-Shot (Top-1)  & -         &5.70$\pm$0.34              & 9.31$\pm$0.43                 & 9.44$\pm$0.43    \\ 
\hhline{~~|-|-|-|-|-}
                                & &  SBERT-InterpEVAL  & -              & 15.55$\pm$0.53                 & 16.31$\pm$0.54                & 13.07$\pm$0.50  \\ 
\hhline{~~|-|-|-|-|-}
                                & & Zero-Shot (Top-5)           & -     & 14.30$\pm$0.52               & 19.81$\pm$0.59                & 19.41$\pm$0.58      \\ 
\hhline{~~|-|-|-|-|-}
                           & &     \begin{tabular}[c]{@{}c@{}}Emb. Extract. (768 dim.)\end{tabular}  & \begin{tabular}[c]{@{}c@{}} RBF-SVC (C=100)\end{tabular}                      & 59.77$\pm$0.72     & 58.82$\pm$0.72      & 39.30$\pm$0.72      \\
\hline
\multirow{4}{*}{Llama-2}         &\multirow{4}{*}{7B}& Zero-Shot          & -  & 29.25$\pm$0.67                & 42.55$\pm$0.73                 & 40.38$\pm$0.72    \\ 
\hhline{~~|-|-|-|-|-}
                                & &  SBERT-InterpEVAL   & -             & 33.44$\pm$0.69                 & 37.32$\pm$0.71                & 27.54$\pm$0.66  \\ 
\hhline{~~|-|-|-|-|-}

                                & & N-Shots     & -         & 33.29$\pm$0.69                & 47.29$\pm$0.74               & 45.05$\pm$0.73    \\ 
\hhline{~~|-|-|-|-|-}
                           & &     \begin{tabular}[c]{@{}c@{}}Emb. Extract. (4096 dim.)\end{tabular}  & \begin{tabular}[c]{@{}c@{}} Lin.-SVC (C=0.001)\end{tabular}                      & \textbf{65.00$\pm$0.70*}     & \textbf{63.75$\pm$0.71}      & 46.21$\pm$0.73      \\
\hline
\multirow{4}{*}{MentaLlama}     &\multirow{4}{*}{7B} & Zero-Shot  & -          & 16.81$\pm$0.55               & 25.95$\pm$0.65                 & 28.46$\pm$0.66     \\ 
\hhline{~~|-|-|-|-|-}
                                & &  SBERT-InterpEVAL & -               & 34.58$\pm$0.70                 & 37.14$\pm$0.71                & 27.64$\pm$0.66  \\ 
\hhline{~~|-|-|-|-|-}
                                & & N-Shots    & -          & 31.09$\pm$0.68               & 43.85$\pm$0.73                & \textbf{49.28$\pm$0.74*}     \\
\hhline{~~|-|-|-|-|-}
                           & &     \begin{tabular}[c]{@{}c@{}}Emb. Extract. (4096 dim.)\end{tabular}  & \begin{tabular}[c]{@{}c@{}} Lin.-SVC (C=0.001)\end{tabular}                      & \textbf{64.61$\pm$0.70}     & \textbf{63.24$\pm$0.71}      & 43.73$\pm$0.73      \\
\hline
\end{tabular}
\end{table*}

\subsection{RQ2. Comparison between LMs: RoBERTa vs. Llama}
Moving to the analysis of differences between architectures, the Llama2-family always surpasses those based on RoBERTa on W-F1 and M-F1 when we perform in-domain comparisons (i.e. the two mental-health adapted models and the two general domain models). 

Specifically for the general domain, comparing the SBERT-InterpEVAL method of RoBERTa and Llama-2, Llama-2 beats RoBERTa achieving a statistically significant increment of 15.55 and 12.83 percentage points on W-F1 and M-F1, respectively, in the Counsel-Chat dataset (Table \ref{tab:ZeroShotTopic}). For 7Cups, the increments are 9.28 and 6.38 for W-F1 and M-F1, respectively.
For the mental-health models, MentaLlama achieves an increment of 19.26 in W-F1 and 18.17 in M-F1 percentage points compared to Mental-RoBERTa in Counsel-Chat. For 7 cups the increment achieved by MentaLlama reaches 18.57 percentage points in W-F1 and 12.47 for M-F1. The same conclusion can be extracted by comparing embedding extraction strategies for both datasets and across the model's architectures.

Nevertheless, to conclude this analysis between families of LMs, it is necessary to mention that Llama-2 models contain a larger number of parameters compared to RoBERTa models, and although in these experiments quantization of bitsandbytes libraries was employed to compress the size of Llama-2, the processing time and the GPU memory were higher for Llama compared to RoBERTa. Equivalently, the size of the embeddings extracted from Llama-2 (4,096 dimensional) is 4 times bigger compared to those generated from RoBERTa (768-dimensional) which could still be a reasonable solution in limited resources scenarios.


\subsection{RQ3. Comparison between Zero-Shot approaches}
Focusing first on the Zero-Shot strategies of RoBERTa and Mental-RoBERTa models, `Zero-Shot (Top-5)' achieved the highest performance compared to the regular version of `Top-1' for both datasets. This result indicates that despite 1) the relationship between topics, 2) the lack of adaptation (or fine-tuning) to the task, and 3) the existence of possible lexical mismatches, these models can differentiate between fine Grained classes on their 5-top predictions (W-F1 of 32.23 for RoBERTa and 22.28 for Mental-RoBERTa on the Counsel-Chat dataset; W-F1 of 35.02 for RoBERTa and 25.83 for Mental-RoBERTa on the 7Cups dataset).

However, the top-1 and Zero-Shot (Top-5) strategies have the limitations of searching for specific matches, e.g. if the top-1 prediction was `anger' and the label of the topic was `anger-management', this result is counted as an error, which could give a false perception of the actual capabilities of these models. To create fairer comparisons with Llama-versions, we included the SBERT-InterpEVAL method, which exploits semantic distances between the top-1 prediction and the list of possible topics. As we can see in the rows of the SBERT-InterpEVAL strategy of Table \ref{tab:ZeroShotTopic} for Counsel-Chat, with this method all the metrics increase in the range of [7.49-14.63] percentage points compared to Zero-Shot (top-1), which indicates that evaluating models only applying Zero-Shots is likely underestimating their capacities, which is later confirmed with the experiments of embeddings' extraction for both datasets. Still, if it is necessary to perform Zero-Shot, SBERT-InterpEVAL is the best strategy for returning single topics from a provided set since it permits recovering from synonyms and lemmatization errors.

\subsection{Comparison with baseline and previous works}
From the comparison of embedding extraction methodologies, we can see that the vectors from RoBERTa, Mental-RoBERTa, Llama-2 and MentaLlama surpass the Word2Vec baseline for both datasets, which encourages its preferential use against this technique. Moving to the comparison between the SoTa SBERT and the embeddings extracted from the evaluated Language Models, we can observe from the results in bold of Tables \ref{tab:ZeroShotTopic} and \ref{tab:ZeroShotTopic7cups} that the embeddings of Llama-2 and MentaLlama obtain comparable results from a statistical perspective to those achieved by SBERT, which is remarkable considering that these models were not initially aimed to solve classification tasks, nevertheless, their larger number of parameters and datasets employed to train them becomes them into a competitive multi-task solution.


\section{Conclusion}\label{sec:conclusions}

Mental health is a major issue in our society, and the lack of resources dedicated to solving this situation enforces the development of technical solutions to assist the increasing number of help-seekers requiring assistance. However, high-quality datasets in the mental health domain for training these systems are still quite limited. Moreover, novel LM models are starting to appear in the market as alternative solutions to rule-based conversational agents. Due to this change of paradigm, it is a prior requirement to analyze their capabilities in different scenarios, dedicating special attention to sensitive domains such as mental health, since open models can be employed by any user independently of their conditions. For this reason, in this article, efforts are dedicated, first, to acquiring a large dataset from a platform moderated by expert therapists with real-world posts; and second, to compare state-of-the-art solutions for performing fine-grain topic classification in the mental-health domain to address three relevant research questions. 

Regarding the first contribution, we acquired a dataset (7Cups) with more than 142,000 samples captured in a real-world scenario. These samples are extracted from the forum's threads regulated by expert therapists, and they are constituted by: 1) questions posted by people possibly suffering from a mental health disorder requiring assistance, and 2) answers given by experts or qualified participants. All the questions belong to a specific topic related to a mental health disease or its possible causes. This dataset is employed together with the CounselChat dataset to analyze the generalization ability of the evaluated models.

As stated before the reduced number of high-quality and large datasets is still a concern in this domain. For this reason, the second contribution of this article is to explore the differences of novel LMs, some of them adapted with mental-health annotated data; hence, potentially aimed to be employed in this domain. Specifically, in the article we gave answers to three research questions in this regard:

1) RQ1 aspirated to solve whether mental-health adapted models could outperform their open-domain versions, for solving fine-grained mental-health topic classification. Their performance on this task was considered a minimum pre-requisite that LLM models should accomplish for maintaining a conversation with a person suffering from a mental health disorder since the first step for participating in a conversation is to understand the context of the dialogue. Counter-intuitively, adapted models to the mental health domain did not always perform better than their open-domain homologous in Zero-Shot scenarios which is not replicated for the embedding extraction strategies in which adapted models seem to be superior in M-F1, i.e. in classifying fine-grain classes with not many representative samples. These outcomes highlight that there are statistically significant differences between the conclusions that can be reached from Zero-Shot strategies compared to embeddings extraction, where the Zero-Shot strategies are dependent on the design of prompts which may result in more pronounced differences; and that those models adapted to mental-health seems not to be as impactful as initially expected, maybe because they used data extracted from social networks with a more informal tone in comparison to the employed in Counsel-Chat or 7Cups. This last observation motivates the pursuit of the question: what does `a similar domain' mean?

2) The RQ2 aimed to evaluate two novel Language Models with distinct architectures, RoBERTa and Llama-2. The results from the Zero-Shot and embedding extraction experiments demonstrate that Llama-2 based models achieve higher recognition rates in most situations, even though this architecture was not initially devised to solve classification but to generate responses as a conversational agent. In addition, its larger size in terms of parameters compared to RoBERTa (70 times smaller) and the amount of data employed for training Llama-2, positions this model as the most competitive in terms of performance. Still, it is important to remark that Llama-2 models require higher computation capabilities than RoBERTa, even when applying compression and double-quantization on the parameters, which is a limiting factor in some scenarios. 

3) The RQ3 focused on the development of a strategy for comparing Language Models' responses in Zero-Shot scenarios since this is one of the most employed techniques to evaluate models recently in the literature, however, there exists no exact match between Zero-Shot strategies when different families of models needs to be appraised in a specific task. To solve this gap, we propose the SBERT-InterpEVAL method. For the first smaller dataset (`Counsel-Chat'), SBERT-InterpEVAL achieved an improvement of 20.12 percentage points in accuracy compared to the Zero-Shot (Top-1) version of RoBERTa, and a 5.75 for the Zero-Shot version of Llama-2. For the second dataset of 7-Cups, it achieved ACC improvements of 14.34 for RoBERTa and 4.19 percentage points for Llama-2, respectively. Apart from the improvements in classification metrics, this methodology also offers the advantages of permitting fine-grain error analysis across classes, and comparisons between different families of transformers as they are RoBERTa and LLama-2 in fairer and more objective conditions.

In future work, we would like to continue deepening into the understanding of the knowledge embedded in state-of-the-art solutions for other domains, continuing the study of fine-tuning from the perspective of required resources for convergence, similarity between domains, and effectiveness in adaptation compared to the open-domain versions of models. 

\section{Limitations}
As datasets are collected from real-world forums for helping users potentially suffering from mental health disorders, some labels could be noisy since users decide where to post their questions. Considering that users are searching for help and assistance, they might be unaware of their condition. 



Another limitation of the SBERT-InterpEVAL approach is that it depends on another model, in our case SBERT. Since a sequential pipeline always drags errors between stages, the top performance of SBERT-InterpEVAL is aligned with the performance of SBERT. For this reason, we selected a state-of-the-art model for detecting textual similarities. This methodology allows us to evaluate how multi-task models (RoBERTa or Llama-2) can work in tailored tasks without adaptation. 

Finally, the prompt itself could also have limitations because there could exist a prompt that obtains higher scores; however, this drawback is partially solved by the incorporation of the embeddings-based strategies that help to extract more robust conclusions about differences in models.  

To conclude, as the dataset employed was mostly in English, no other languages are represented. Neither the current investigation covers all demographic groups or all the topics that may appear in the mental health domain.

\section{Ethical Aspects}

Models and data generated from this study were not aimed to be employed directly with users suffering from severe mental health issues, but to advance the scientific knowledge on the capacities and limitations of language models for handling mental-health-related topics, since currently these models are starting to be employed in many domains without attending to their risks. Indeed, this is one of the main motivations of this study: to encourage performing fine-grained evaluations on different fields to understand and tackle their limitations.

\section{CRediT Contributions}

Software: C.L.-J.;
Data curation: C.L.-J.;
Visualization: C.L.-J.;
Formal analysis: C.L.-J.;
Validation: C.L.-J.;
Writing - original draft: C.L.-J.,N.W., D.G., Z.C.,S.U.;
Investigation: C.L.-J., D.G., Z.C., N.W., E.A., S.U.;
Conceptualization: C.L.-J., D.G., Z.C.,N.W.;
Methodology: C.L.-J., 
Supervision:  D.G., Z.C., E.A.,S.U.;
Funding acquisition:  D.G., Z.C., E.A.,S.U.

\section*{Acknowledgements}

The research leading to these results has received funding from the Spanish Ministry of Science and Innovation through the projects  GOMINOLA (PID2020-118112RB-C21 and PID2020-118112RB-C22) financed by \\ MCIN/AEI/10.13039/501100011033,  and TRUSTBOOST (PID2023-150584OB-C22). The TrustBoost project has received funding from MICIU/AEI/10.13039/501100011033 and from FEDER, UE. It has also received funding from the project FORSocialRobots, financed by the Bavarian Research Foundation (AZ1594-23).

The founder played no role in study design, data collection, analysis and interpretation of data, or the writing of this manuscript.

\bibliographystyle{model5-names}
\bibliography{references}

@misc{squizofreniaQA,
      title={Question-Answering Model for Schizophrenia Symptoms and Their Impact on Daily Life using Mental Health Forums Data}, 
      author={Christian Internò and Eloisa Ambrosini},
      year={2023},
      eprint={2310.00448},
      archivePrefix={arXiv},
      primaryClass={cs.LG}
}

@article{MentalHealthAgent,
author = {Eagle, Tessa and Blau, Conrad and Bales, Sophie and Desai, Noopur and Li, Victor and Whittaker, Steve},
title = {“I don’t know what you mean by `I am anxious'”: A New Method for Evaluating Conversational Agent Responses to Standardized Mental Health Inputs for Anxiety and Depression},
year = {2022},
issue_date = {June 2022},
publisher = {Association for Computing Machinery},
address = {New York, NY, USA},
volume = {12},
number = {2},
issn = {2160-6455},
url = {https://doi.org/10.1145/3488057},
doi = {10.1145/3488057},
journal = {ACM Trans. Interact. Intell. Syst.},
month = {jul},
articleno = {12},
numpages = {23}
}

@InProceedings{iwinac2024,
author="Luna-Jim{\'e}nez, Cristina
and Griol, David
and Callejas, Zoraida",
editor="Ferr{\'a}ndez Vicente, Jos{\'e} Manuel
and Val Calvo, Mikel
and Adeli, Hojjat",
title="Zero-Shot Ensemble of Language Models for Fine-Grain Mental-Health Topic Classification",
booktitle="Artificial Intelligence for Neuroscience and Emotional Systems",
year="2024",
publisher="Springer Nature Switzerland",
address="Cham",
pages="88--97",
isbn="978-3-031-61140-7"
}

@INPROCEEDINGS {cbms24,
author = {C. Luna-Jiménez, Z. Callejas and D. Griol},
booktitle = {(Accepted) 2024 IEEE 37th International Symposium on Computer-Based Medical Systems (CBMS)},
title = {Mental-Health Topic Classification employing D-vectors of Large Language Models},
year = {2024},
volume = {},
issn = {},
pages = {},
publisher = {IEEE Computer Society},
address = {Los Alamitos, CA, USA},
month = {}
}

@misc{bertagnolli2020counsel,
  title={Counsel chat: Bootstrapping high-quality therapy data},
  author={Bertagnolli, Nicolas},
  year={2020},
  publisher={Towards Data Science},
  url= {https://towardsdatascience.com/counsel-chat-bootstrapping-high-quality-therapy-data-971b419f33da}
}

@misc{WHO_mentalHealth,
  title={World mental health report: transforming mental health for all.},
author = {World Health Organization},
  year={2022},
  publisher={Geneva: World Health Organization},
 url={https://iris.who.int/bitstream/handle/10665/356119/9789240049338-eng.pdf?isAllowed=y&sequence=1}
}

@article{RodrguezCantelar2023,
  title = {Automatic Detection of Inconsistencies and Hierarchical Topic Classification for Open-Domain Chatbots},
  volume = {13},
  ISSN = {2076-3417},
  url = {http://dx.doi.org/10.3390/app13169055},
  DOI = {10.3390/app13169055},
  number = {16},
  journal = {Applied Sciences},
  publisher = {MDPI AG},
  author = {Rodríguez-Cantelar,  Mario and Estecha-Garitagoitia,  Marcos and D’Haro,  Luis Fernando and Matía,  Fernando and Córdoba,  Ricardo},
  year = {2023},
  month = aug,
  pages = {9055}
}

@article{yang2023mentalllama,
  title={MentalLLaMA: Interpretable Mental Health Analysis on Social Media with Large Language Models},
  author={Yang, Kailai and Zhang, Tianlin and Kuang, Ziyan and Xie, Qianqian and Ananiadou, Sophia},
  journal={arXiv preprint arXiv:2309.13567},
  year={2023}
}

@inproceedings{sarkar-etal-2023-zero,
    title = "Zero-Shot Multi-Label Topic Inference with Sentence Encoders and {LLM}s",
    author = "Sarkar, Souvika  and
      Feng, Dongji  and
      Karmaker Santu, Shubhra Kanti",
    editor = "Bouamor, Houda  and
      Pino, Juan  and
      Bali, Kalika",
    booktitle = "Proceedings of the 2023 Conference on Empirical Methods in Natural Language Processing",
    month = dec,
    year = "2023",
    address = "Singapore",
    publisher = "Association for Computational Linguistics",
    url = "https://aclanthology.org/2023.emnlp-main.1008",
    doi = "10.18653/v1/2023.emnlp-main.1008",
    pages = "16218--16233",
}

@INPROCEEDINGS {xception,
author = {F. Chollet},
booktitle = {2017 IEEE Conference on Computer Vision and Pattern Recognition (CVPR)},
title = {Xception: Deep Learning with Depthwise Separable Convolutions},
year = {2017},
volume = {},
issn = {1063-6919},
pages = {1800-1807},
doi = {10.1109/CVPR.2017.195},
url = {https://doi.ieeecomputersociety.org/10.1109/CVPR.2017.195},
publisher = {IEEE Computer Society},
address = {Los Alamitos, CA, USA},
month = {jul}
}

@inproceedings{yang-etal-2023-towards,
    title = "Towards Interpretable Mental Health Analysis with Large Language Models",
    author = "Yang, Kailai  and
      Ji, Shaoxiong  and
      Zhang, Tianlin  and
      Xie, Qianqian  and
      Kuang, Ziyan  and
      Ananiadou, Sophia",
    editor = "Bouamor, Houda  and
      Pino, Juan  and
      Bali, Kalika",
    booktitle = "Proceedings of the 2023 Conference on Empirical Methods in Natural Language Processing",
    month = dec,
    year = "2023",
    address = "Singapore",
    publisher = "Association for Computational Linguistics",
    url = "https://aclanthology.org/2023.emnlp-main.370",
    doi = "10.18653/v1/2023.emnlp-main.370",
    pages = "6056--6077",
}

@inproceedings{sbert,
    title = "Sentence-{BERT}: Sentence Embeddings using {S}iamese {BERT}-Networks",
    author = "Reimers, Nils  and
      Gurevych, Iryna",
    editor = "Inui, Kentaro  and
      Jiang, Jing  and
      Ng, Vincent  and
      Wan, Xiaojun",
    booktitle = "Proceedings of the 2019 Conference on Empirical Methods in Natural Language Processing and the 9th International Joint Conference on Natural Language Processing (EMNLP-IJCNLP)",
    month = nov,
    year = "2019",
    address = "Hong Kong, China",
    publisher = "Association for Computational Linguistics",
    url = "https://aclanthology.org/D19-1410",
    doi = "10.18653/v1/D19-1410",
    pages = "3982--3992",
}

@InProceedings{CLEFconf,
author="Losada, David E.
and Crestani, Fabio",
editor="Fuhr, Norbert
and Quaresma, Paulo
and Gon{\c{c}}alves, Teresa
and Larsen, Birger
and Balog, Krisztian
and Macdonald, Craig
and Cappellato, Linda
and Ferro, Nicola",
title="A Test Collection for Research on Depression and Language Use",
booktitle="Experimental IR Meets Multilinguality, Multimodality, and Interaction",
year="2016",
publisher="Springer International Publishing",
address="Cham",
pages="28--39",
isbn="978-3-319-44564-9"
}

@InProceedings{LTEDI2022,
author="Sampath, Kayalvizhi
and Durairaj, Thenmozhi",
editor="Kalinathan, Lekshmi
and R., Priyadharsini
and Kanmani, Madheswari
and S., Manisha",
title="Data Set Creation and Empirical Analysis for Detecting Signs of Depression from Social Media Postings",
booktitle="Computational Intelligence in Data Science",
year="2022",
publisher="Springer International Publishing",
address="Cham",
pages="136--151",
isbn="978-3-031-16364-7"
}

@inproceedings{CLPsych19,
    title = "{CLP}sych 2019 Shared Task: Predicting the Degree of Suicide Risk in {R}eddit Posts",
    author = {Zirikly, Ayah  and
      Resnik, Philip  and
      Uzuner, {\"O}zlem  and
      Hollingshead, Kristy},
    editor = "Niederhoffer, Kate  and
      Hollingshead, Kristy  and
      Resnik, Philip  and
      Resnik, Rebecca  and
      Loveys, Kate",
    booktitle = "Proceedings of the Sixth Workshop on Computational Linguistics and Clinical Psychology",
    month = jun,
    year = "2019",
    address = "Minneapolis, Minnesota",
    publisher = "Association for Computational Linguistics",
    url = "https://aclanthology.org/W19-3003",
    doi = "10.18653/v1/W19-3003",
    pages = "24--33",
}

@inproceedings{SenseMood,
author = {Lin, Chenhao and Hu, Pengwei and Su, Hui and Li, Shaochun and Mei, Jing and Zhou, Jie and Leung, Henry},
title = {SenseMood: Depression Detection on Social Media},
year = {2020},
isbn = {9781450370875},
publisher = {Association for Computing Machinery},
address = {New York, NY, USA},
doi = {10.1145/3372278.3391932},
booktitle = {Proceedings of the 2020 International Conference on Multimedia Retrieval},
pages = {407–411},
numpages = {5},
location = {Dublin, Ireland},
series = {ICMR '20}
}

@inproceedings{CAMS,
    title = "{CAMS}: An Annotated Corpus for Causal Analysis of Mental Health Issues in Social Media Posts",
    author = "Garg, Muskan  and
      Saxena, Chandni  and
      Saha, Sriparna  and
      Krishnan, Veena  and
      Joshi, Ruchi  and
      Mago, Vijay",
    editor = "Calzolari, Nicoletta  and
      B{\'e}chet, Fr{\'e}d{\'e}ric  and
      Blache, Philippe  and
      Choukri, Khalid  and
      Cieri, Christopher  and
      Declerck, Thierry  and
      Goggi, Sara  and
      Isahara, Hitoshi  and
      Maegaard, Bente  and
      Mariani, Joseph  and
      Mazo, H{\'e}l{\`e}ne  and
      Odijk, Jan  and
      Piperidis, Stelios",
    booktitle = "Proceedings of the Thirteenth Language Resources and Evaluation Conference",
    month = jun,
    year = "2022",
    address = "Marseille, France",
    publisher = "European Language Resources Association",
    url = "https://aclanthology.org/2022.lrec-1.686",
    pages = "6387--6396",
}

@inproceedings{daicWOz,
    title = "The Distress Analysis Interview Corpus of human and computer interviews",
    author = "Gratch, Jonathan  and
      Artstein, Ron  and
      Lucas, Gale  and
      Stratou, Giota  and
      Scherer, Stefan  and
      Nazarian, Angela  and
      Wood, Rachel  and
      Boberg, Jill  and
      DeVault, David  and
      Marsella, Stacy  and
      Traum, David  and
      Rizzo, Skip  and
      Morency, Louis-Philippe",
    editor = "Calzolari, Nicoletta  and
      Choukri, Khalid  and
      Declerck, Thierry  and
      Loftsson, Hrafn  and
      Maegaard, Bente  and
      Mariani, Joseph  and
      Moreno, Asuncion  and
      Odijk, Jan  and
      Piperidis, Stelios",
    booktitle = "Proceedings of the Ninth International Conference on Language Resources and Evaluation ({LREC}'14)",
    month = may,
    year = "2014",
    address = "Reykjavik, Iceland",
    publisher = "European Language Resources Association (ELRA)",
    url = "http://www.lrec-conf.org/proceedings/lrec2014/pdf/508_Paper.pdf",
    pages = "3123--3128",
}

@inproceedings{hopeDS,
author = {Malhotra, Ganeshan and Waheed, Abdul and Srivastava, Aseem and Akhtar, Md Shad and Chakraborty, Tanmoy},
title = {Speaker and Time-Aware Joint Contextual Learning for Dialogue-Act Classification in Counselling Conversations},
year = {2022},
isbn = {9781450391320},
publisher = {Association for Computing Machinery},
address = {New York, NY, USA},
doi = {10.1145/3488560.3498509},
booktitle = {Proceedings of the Fifteenth ACM International Conference on Web Search and Data Mining},
pages = {735–745},
numpages = {11},
location = {Virtual Event, AZ, USA},
series = {WSDM '22}
}

@inproceedings{memoDS,
author = {Srivastava, Aseem and Suresh, Tharun and Lord, Sarah P. and Akhtar, Md Shad and Chakraborty, Tanmoy},
title = {Counseling Summarization Using Mental Health Knowledge Guided Utterance Filtering},
year = {2022},
isbn = {9781450393850},
publisher = {Association for Computing Machinery},
address = {New York, NY, USA},
doi = {10.1145/3534678.3539187},
booktitle = {Proceedings of the 28th ACM SIGKDD Conference on Knowledge Discovery and Data Mining},
pages = {3920–3930},
numpages = {11},
location = {Washington DC, USA},
series = {KDD '22}
}

@article{woebot,
  title = {Delivering Cognitive Behavior Therapy to Young Adults With Symptoms of Depression and Anxiety Using a Fully Automated Conversational Agent (Woebot): A Randomized Controlled Trial},
  volume = {4},
  ISSN = {2368-7959},
  url = {http://dx.doi.org/10.2196/mental.7785},
  
  number = {2},
  journal = {JMIR Mental Health},
  publisher = {JMIR Publications Inc.},
  author = {Fitzpatrick,  Kathleen Kara and Darcy,  Alison and Vierhile,  Molly},
  year = {2017},
  month = jun,
  pages = {e19}
}

@article{wysa,
  title = {An Empathy-Driven,  Conversational Artificial Intelligence Agent (Wysa) for Digital Mental Well-Being: Real-World Data Evaluation Mixed-Methods Study},
  volume = {6},
  ISSN = {2291-5222},
  url = {http://dx.doi.org/10.2196/12106},
    DOI = {10.2196/12106},
  number = {11},
  journal = {JMIR mHealth and uHealth},
  publisher = {JMIR Publications Inc.},
  author = {Inkster,  Becky and Sarda,  Shubhankar and Subramanian,  Vinod},
  year = {2018},
  month = nov,
  pages = {e12106}
}

@InProceedings{Kraus2021,
author="Kraus, Matthias
and Seldschopf, Philip
and Minker, Wolfgang",
editor="Loko{\v{c}}, Jakub
and Skopal, Tom{\'a}{\v{s}}
and Schoeffmann, Klaus
and Mezaris, Vasileios
and Li, Xirong
and Vrochidis, Stefanos
and Patras, Ioannis",
title="Towards the Development of a Trustworthy Chatbot for Mental Health Applications",
booktitle="MultiMedia Modeling",
year="2021",
publisher="Springer International Publishing",
address="Cham",
pages="354--366",
isbn="978-3-030-67835-7"
}

@Inbook{Callejas2021review,
author="Callejas, Zoraida
and Griol, David",
editor="Lopez-Soto, Teresa",
title="Conversational Agents for Mental Health and Wellbeing",
bookTitle="Dialog Systems: A Perspective from Language, Logic and Computation",
year="2021",
publisher="Springer International Publishing",
address="Cham",
pages="219--244",
volume = "22",
isbn="978-3-030-61438-6",
doi="10.1007/978-3-030-61438-6_11",
url="https://doi.org/10.1007/978-3-030-61438-6_11"
}

@inproceedings{gpt3,
author = {Brown, Tom B. and Mann, Benjamin and Ryder, Nick and Subbiah, Melanie and Kaplan, Jared and Dhariwal, Prafulla and Neelakantan, Arvind and Shyam, Pranav and Sastry, Girish and Askell, Amanda and Agarwal, Sandhini and Herbert-Voss, Ariel and Krueger, Gretchen and Henighan, Tom and Child, Rewon and Ramesh, Aditya and Ziegler, Daniel M. and Wu, Jeffrey and Winter, Clemens and Hesse, Christopher and Chen, Mark and Sigler, Eric and Litwin, Mateusz and Gray, Scott and Chess, Benjamin and Clark, Jack and Berner, Christopher and McCandlish, Sam and Radford, Alec and Sutskever, Ilya and Amodei, Dario},
title = {Language models are few-shot learners},
year = {2020},
isbn = {9781713829546},
publisher = {Curran Associates Inc.},
address = {Red Hook, NY, USA},
booktitle = {Proceedings of the 34th International Conference on Neural Information Processing Systems},
articleno = {159},
numpages = {25},
location = {Vancouver, BC, Canada},
series = {NIPS'20}
}

@article{llama2,
      title={Llama 2: Open Foundation and Fine-Tuned Chat Models}, 
      author={Hugo Touvron and Louis Martin and Kevin Stone and Peter Albert and Amjad Almahairi and Yasmine Babaei and Nikolay Bashlykov and Soumya Batra and Prajjwal Bhargava and Shruti Bhosale and Dan Bikel and Lukas Blecher and Cristian Canton Ferrer and Moya Chen and Guillem Cucurull and David Esiobu and Jude Fernandes and Jeremy Fu and Wenyin Fu and Brian Fuller and Cynthia Gao and Vedanuj Goswami and Naman Goyal and Anthony Hartshorn and Saghar Hosseini and Rui Hou and Hakan Inan and Marcin Kardas and Viktor Kerkez and Madian Khabsa and Isabel Kloumann and Artem Korenev and Punit Singh Koura and Marie-Anne Lachaux and Thibaut Lavril and Jenya Lee and Diana Liskovich and Yinghai Lu and Yuning Mao and Xavier Martinet and Todor Mihaylov and Pushkar Mishra and Igor Molybog and Yixin Nie and Andrew Poulton and Jeremy Reizenstein and Rashi Rungta and Kalyan Saladi and Alan Schelten and Ruan Silva and Eric Michael Smith and Ranjan Subramanian and Xiaoqing Ellen Tan and Binh Tang and Ross Taylor and Adina Williams and Jian Xiang Kuan and Puxin Xu and Zheng Yan and Iliyan Zarov and Yuchen Zhang and Angela Fan and Melanie Kambadur and Sharan Narang and Aurelien Rodriguez and Robert Stojnic and Sergey Edunov and Thomas Scialom},
      year={2023},
      eprint={2307.09288},
      archivePrefix={arXiv},
      primaryClass={cs.CL}
}

@article{bertSentiment,
  doi = {10.1186/s40537-023-00781-w},
  url = {https://doi.org/10.1186/s40537-023-00781-w},
  year = {2023},
  month = jun,
  publisher = {Springer Science and Business Media {LLC}},
  volume = {10},
  number = {1},
  author = {Amira Samy Talaat},
  title = {Sentiment analysis classification system using hybrid {BERT} models},
  journal = {Journal of Big Data}
}

@article{BERTemotions, title={BERT-ERC: Fine-Tuning BERT Is Enough for Emotion Recognition in Conversation}, volume={37}, url={https://ojs.aaai.org/index.php/AAAI/article/view/26582}, DOI={10.1609/aaai.v37i11.26582}, abstractNote={Previous works on emotion recognition in conversation (ERC) follow a two-step paradigm, which can be summarized as first producing context-independent features via fine-tuning pretrained language models (PLMs) and then analyzing contextual information and dialogue structure information among the extracted features. However, we discover that this paradigm has several limitations. Accordingly, we propose a novel paradigm, i.e., exploring contextual information and dialogue structure information in the fine-tuning step, and adapting the PLM to the ERC task in terms of input text, classification structure, and training strategy. Furthermore, we develop our model BERT-ERC according to the proposed paradigm, which improves ERC performance in three aspects, namely suggestive text, fine-grained classification module, and two-stage training. Compared to existing methods, BERT-ERC achieves substantial improvement on four datasets, indicating its effectiveness and generalization capability. Besides, we also set up the limited resources scenario and the online prediction scenario to approximate real-world scenarios. Extensive experiments demonstrate that the proposed paradigm significantly outperforms the previous one and can be adapted to various scenes.}, number={11}, journal={Proceedings of the AAAI Conference on Artificial Intelligence}, author={Qin, Xiangyu and Wu, Zhiyu and Zhang, Tingting and Li, Yanran and Luan, Jian and Wang, Bin and Wang, Li and Cui, Jinshi}, year={2023}, month={Jun.}, pages={13492-13500} }

@inproceedings{BERT,
    title = "{BERT}: Pre-training of Deep Bidirectional Transformers for Language Understanding",
    author = "Devlin, Jacob  and
      Chang, Ming-Wei  and
      Lee, Kenton  and
      Toutanova, Kristina",
    booktitle = "Proceedings of the 2019 Conference of the North {A}merican Chapter of the Association for Computational Linguistics: Human Language Technologies, Volume 1 (Long and Short Papers)",
    month = jun,
    year = "2019",
    address = "Minneapolis, Minnesota",
    publisher = "Association for Computational Linguistics",
    url = "https://aclanthology.org/N19-1423",
    doi = "10.18653/v1/N19-1423",
    pages = "4171--4186",
}

@inproceedings{mentalbert,
    title = "{M}ental{BERT}: Publicly Available Pretrained Language Models for Mental Healthcare",
    author = "Ji, Shaoxiong  and
      Zhang, Tianlin  and
      Ansari, Luna  and
      Fu, Jie  and
      Tiwari, Prayag  and
      Cambria, Erik",
    booktitle = "Proceedings of the Thirteenth Language Resources and Evaluation Conference",
    month = jun,
    year = "2022",
    address = "Marseille, France",
    publisher = "European Language Resources Association",
    url = "https://aclanthology.org/2022.lrec-1.778",
    pages = "7184--7190",
}

@inproceedings{transformer,
author = {Vaswani, Ashish and Shazeer, Noam and Parmar, Niki and Uszkoreit, Jakob and Jones, Llion and Gomez, Aidan N. and Kaiser, \L{}ukasz and Polosukhin, Illia},
title = {Attention is All You Need},
year = {2017},
isbn = {9781510860964},
publisher = {Curran Associates Inc.},
address = {Red Hook, NY, USA},
booktitle = {Proceedings of the 31st International Conference on Neural Information Processing Systems},
pages = {6000–6010},
numpages = {11},
location = {Long Beach, California, USA},
series = {NIPS'17}
}

@inproceedings{carolineArgSys,
    title = "From Argument Search to Argumentative Dialogue: A Topic-independent Approach to Argument Acquisition for Dialogue Systems",
    author = "Rach, Niklas  and
      Schindler, Carolin  and
      Feustel, Isabel  and
      Daxenberger, Johannes  and
      Minker, Wolfgang  and
      Ultes, Stefan",
    editor = "Li, Haizhou  and
      Levow, Gina-Anne  and
      Yu, Zhou  and
      Gupta, Chitralekha  and
      Sisman, Berrak  and
      Cai, Siqi  and
      Vandyke, David  and
      Dethlefs, Nina  and
      Wu, Yan  and
      Li, Junyi Jessy",
    booktitle = "Proceedings of the 22nd Annual Meeting of the Special Interest Group on Discourse and Dialogue",
    month = jul,
    year = "2021",
    address = "Singapore and Online",
    publisher = "Association for Computational Linguistics",
    url = "https://aclanthology.org/2021.sigdial-1.39",
    doi = "10.18653/v1/2021.sigdial-1.39",
    pages = "368--379",
}

@article{bertTopic,
  title={BERTopic: Neural topic modeling with a class-based TF-IDF procedure},
  author={Grootendorst, Maarten},
  journal={arXiv preprint arXiv:2203.05794},
  year={2022}
}

@article{bertTopicArabic,
title = {BERT for Arabic Topic Modeling: An Experimental Study on BERTopic Technique},
journal = {Procedia Computer Science},
volume = {189},
pages = {191-194},
year = {2021},
note = {AI in Computational Linguistics},
issn = {1877-0509},
doi = {https://doi.org/10.1016/j.procs.2021.05.096},
url = {https://www.sciencedirect.com/science/article/pii/S1877050921012199},
author = {Abeer Abuzayed and Hend Al-Khalifa}
}

@article{memText,
  title = {Topic-Oriented Text Features Can Match Visual Deep Models of Video Memorability},
  volume = {11},
  ISSN = {2076-3417},
  url = {http://dx.doi.org/10.3390/app11167406},
  DOI = {10.3390/app11167406},
  number = {16},
  journal = {Applied Sciences},
  publisher = {MDPI AG},
  author = {Kleinlein,  Ricardo and Luna-Jiménez,  Cristina and Arias-Cuadrado,  David and Ferreiros,  Javier and Fernández-Martínez,  Fernando},
  year = {2021},
  month = aug,
  pages = {7406}
}

@ARTICLE{zeroShotTopic,
  author={Pourpanah, Farhad and Abdar, Moloud and Luo, Yuxuan and Zhou, Xinlei and Wang, Ran and Lim, Chee Peng and Wang, Xi-Zhao and Wu, Q. M. Jonathan},
  journal={IEEE Transactions on Pattern Analysis and Machine Intelligence}, 
  title={A Review of Generalized Zero-Shot Learning Methods}, 
  year={2023},
  volume={45},
  number={4},
  pages={4051-4070},
  doi={10.1109/TPAMI.2022.3191696}}

@INPROCEEDINGS{facenet,
  author={Schroff, Florian and Kalenichenko, Dmitry and Philbin, James},
  booktitle={2015 IEEE Conference on Computer Vision and Pattern Recognition (CVPR)}, 
  title={FaceNet: A unified embedding for face recognition and clustering}, 
  year={2015},
  volume={},
  number={},
  pages={815-823},
  doi={10.1109/CVPR.2015.7298682}}

@INPROCEEDINGS{topicsModel,
  author={Suryadjaja, Paulus Setiawan and Mandala, Rila},
  booktitle={2021 8th International Conference on Advanced Informatics: Concepts, Theory and Applications (ICAICTA)}, 
  title={Improving the Performance of the Extractive Text Summarization by a Novel Topic Modeling and Sentence Embedding Technique using SBERT}, 
  year={2021},
  volume={},
  number={},
  pages={1-6},
  doi={10.1109/ICAICTA53211.2021.9640295}}

@article{chatPAL,
author="Potts, Courtney
and Lindstr{\"o}m, Frida
and Bond, Raymond
and Mulvenna, Maurice
and Booth, Frederick
and Ennis, Edel
and Parding, Karolina
and Kostenius, Catrine
and Broderick, Thomas
and Boyd, Kyle
and Vartiainen, Anna-Kaisa
and Nieminen, Heidi
and Burns, Con
and Bickerdike, Andrea
and Kuosmanen, Lauri
and Dhanapala, Indika
and Vakaloudis, Alex
and Cahill, Brian
and MacInnes, Marion
and Malcolm, Martin
and O'Neill, Siobhan",
title="A Multilingual Digital Mental Health and Well-Being Chatbot (ChatPal): Pre-Post Multicenter Intervention Study",
journal="J Med Internet Res",
year="2023",
month="Jul",
day="6",
volume="25",
pages="e43051",
issn="1438-8871",
doi="10.2196/43051",
url="https://www.jmir.org/2023/1/e43051",
}

\end{document}